\documentclass[conference]{IEEEtran}
\IEEEoverridecommandlockouts
\usepackage{cite}
\usepackage{amsmath,amssymb,amsfonts}
\usepackage{algorithmic}
\usepackage{graphicx}
\usepackage{textcomp}
\usepackage{xcolor}
\usepackage{multirow}
\usepackage{tabularx}
\usepackage{booktabs}
\usepackage{threeparttable}

\usepackage{multirow}
\usepackage{algorithm}
\usepackage{hyperref}
\usepackage{cleveref}

\begin{document}


\title{Head-Tail-Aware KL Divergence in Knowledge Distillation for Spiking Neural Networks\\
}

\author{
    \IEEEauthorblockN{
        Tianqing Zhang\textsuperscript{\rm 1}\IEEEauthorrefmark{1}, 
        Zixin Zhu\textsuperscript{\rm 2}\IEEEauthorrefmark{1}\thanks{\IEEEauthorrefmark{1}Equal contribution.}, 
        Kairong Yu\textsuperscript{\rm 1}, 
        and Hongwei Wang\textsuperscript{\rm 1,2}\IEEEauthorrefmark{2}\thanks{\IEEEauthorrefmark{2}Corresponding author.}
        }
    \IEEEauthorblockA{
        \textsuperscript{\rm 1}
        College of Computer Science and Technology, Zhejiang University, Hangzhou, China
        } 
    \IEEEauthorblockA{
        \textsuperscript{\rm 2}
        ZJU-UIUC Institute, Zhejiang University, Haining, China
        }
    {\tt\small \textsuperscript{\rm 1}\{zhangtianqing,yukairong\}@zju.edu.cn}
    \hspace{5pt}
    {\tt\small \textsuperscript{\rm 2}\{zixinz.21,hongweiwang\}@intl.zju.edu.cn}
}

\maketitle

\begin{abstract}
Spiking Neural Networks (SNNs) have emerged as a promising approach for energy-efficient and biologically plausible computation. However, due to limitations in existing training methods and inherent model constraints, SNNs often exhibit a performance gap when compared to Artificial Neural Networks (ANNs). Knowledge distillation (KD) has been explored as a technique to transfer knowledge from ANN teacher models to SNN student models to mitigate this gap. Traditional KD methods typically use Kullback-Leibler (KL) divergence to align output distributions. However, conventional KL-based approaches fail to fully exploit the unique characteristics of SNNs, as they tend to overemphasize high-probability predictions while neglecting low-probability ones, leading to suboptimal generalization.
To address this, we propose Head-Tail Aware Kullback-Leibler (HTA-KL) divergence, a novel KD method for SNNs. HTA-KL introduces a cumulative probability-based mask to dynamically distinguish between high- and low-probability regions. It assigns adaptive weights to ensure balanced knowledge transfer, enhancing the overall performance. By integrating forward KL (FKL) and reverse KL (RKL) divergence, our method effectively align both head and tail regions of the distribution. 
We evaluate our methods on CIFAR-10, CIFAR-100 and Tiny ImageNet datasets. Our method outperforms existing methods on most datasets with fewer timesteps.
\end{abstract}


\section{Introduction}
\label{sec:intro}

Spiking Neural Networks (SNNs), regarded as the third generation of neural networks, offer a biologically inspired framework for information processing in neural networks. They transmit information via discrete spikes, enabling inherently low-latency and energy-efficient computation. This makes SNNs particularly well-suited for deployment on neuromorphic computing hardware platforms\cite{maass_networks_1997, yamazaki_spiking_2022, taherkhani_review_2020,guan_relative_2022,liu2024optical,liu2024line,liu2025stereo}.
Despite their biological plausibility and energy efficiency, SNNs currently underperform compared to traditional Artificial Neural Networks (ANNs) in tasks such as image classification, object segmentation, and natural language processing. 
Due to the non-continuous and non-differentiable nature of spike-based activation functions, SNNs cannot be directly trained using the standard Backpropagation (BP) algorithm widely used in ANNs. Instead, they rely on alternative training techniques such as Spatio-Temporal Backpropagation (STBP) or Surrogate Gradient(SG)\cite{wu_spatio-temporal_2018,neftci_surrogate_2019}.

To mitigate the performance gap between ANNs and SNNs, Knowledge Distillation (KD)\cite{huang_knowledge_2022, zhao_decoupled_2022} has been introduced into SNN training. In SNN-based KD, the ANN acts as a teacher model while the SNN serves as the student, enabling the SNN to inherit feature representations from the pretrained ANN while retaining its low-power advantage. Recent works such as KDSNN\cite{xu_constructing_2023}, LaSNN~\cite{hong_lasnn_2023}, and BKDSNN~\cite{xu_bkdsnn_2025} have made significant progress, demonstrating the effectiveness of KD in SNN training.
Given that the primary advantage of SNNs lies in energy efficiency, and that the number of timesteps significantly affects energy consumption, it is crucial to achieve strong performance using as few timesteps as possible. 
Therefore, we aim to develop a more suitable method for applying KD to SNNs that performs better under small timesteps. 
The Kullback-Leibler (KL) loss plays a crucial role in the KD process, and we seek to optimize the current KL function used in SNN knowledge distillation.

Previous distillation studies \cite{wu_rethinking_2025} have found that, during the early training of large language models (LLMs), the forward KL divergence (FKL) tends to align the teacher’s high-probability (head) regions, while Reverse Kullback-Leibler (RKL) divergence focuses more on low-probability (tail) regions.
Based on this observation, a adaptive weighting mechanism called Adaptive KL (AKL) was proposed to combine FKL and RKL, enabling a more comprehensive capture of the teacher model’s distribution within limited training epochs and significantly improving distillation performance.

In this study, we extend the head-tail distribution alignment insight to the SNN setting and propose a novel Head-Tail-Aware KL Divergence (HTA-KL). 
HTA-KL first maps the spike firing rates of the student model over timesteps into a probability distribution, and then adaptively adjusts the weights of the FKL and RKL components based on the alignment difference between the teacher and student models. This allows the SNN to align with both high- and low-probability regions of the distribution without changing the original network architecture or training process.
Moreover, HTA-KL takes into account the unique spatio-temporal dynamics of SNNs, enabling more effective knowledge transfer and improving overall training efficiency.
We validate HTA-KL on CIFAR-10, CIFAR-100, and Tiny ImageNet datasets, achieving superior performance under low timestep conditions. We also conduct extensive ablation studies and analyze the method’s performance on spike firing rates and energy consumption.
Our main contributions are as follows:

\begin{itemize}

\item We propose Head-Tail-Aware KL Divergence (HTA-KL), a novel knowledge distillation framework for SNNs that combines Forward KL (FKL) and Reverse KL (RKL) divergences, enabling SNNs to learn both “head” (high-probability) and “tail” (low-probability) information from the teacher model’s distribution.

\item We introduce an adaptive weighting mechanism that dynamically balances FKL and RKL based on the distribution alignment difference between the teacher and student models, thereby enhancing the transfer of both head and tail distributed knowledge.

\item We conduct comprehensive experiments on CIFAR-10, CIFAR-100, and Tiny ImageNet, along with detailed ablation studies and analyses of spike firing rate and energy efficiency.
\end{itemize}

\section{Related Works}
\label{sec:related_work}

\subsection{SNN Training Methods}


SNN training is primarily categorized into two approaches: ANN-to-SNN conversion and direct training of SNNs. 
The conversion-based approach transfers knowledge from pre-trained ANNs to SNNs by substituting ReLU activations with spike-based neuronal models. 
However, such methods often fall short in fully leveraging the spatio-temporal dynamics inherent to SNNs, limiting their potential in tasks where temporal encoding is critical\cite{wang_new_2023,hu_advancing_2024}.
Direct SNN training, which uses spike-based temporal information, avoids these limitations\cite{zhang_da-lif_2025}. 
Innovations like tdBN\cite{zheng_going_2021} and MPBN\cite{guo_membrane_2023} improve training by optimizing feature normalization and stabilizing membrane potential updates. Frequency-based attention mechanisms have been introduced to optimize intermediate spike features and reduce redundant spiking\cite{yu_fsta-snnfrequency-based_2025}.
ESL-SNNs\cite{shen_esl-snns_2023} propose an efficient evolutionary structure learning framework for SNNs.
Several modifications to ResNet architectures, including SEW-ResNet\cite{fang_deep_2021}, DS-ResNet\cite{feng_multi-level_2022}, and MS-ResNet \cite{hu_advancing_2024}, have also been explored to enhance performance. 
Additionally, networks like RSNN\cite{xu_rsnn_2024} enhances spatiotemporal processing by using a recurrent structure to preprocess input slices.
The integration of Transformer-based SNNs is another area of focus, with models like\cite{shi_spikingresformer_2024,zhou_spikformer_2022,shen_temporal_2025} improving classification accuracy using spike-based self-attention mechanisms. 
STMixer\cite{deng_spiking_2024} models is designed for event-driven asynchronous chips.

\subsection{Knowledge Distillation for SNNs}

KD is a widely adopted model compression method, where a smaller student model is trained to replicate the behavior of a larger teacher model. This strategy allows the student model to maintain competitive accuracy while significantly reducing computational cost and model size\cite{yuan_revisiting_2020,chen_distilling_2021,sun_logit_2024}.
As a critical approach for enhancing performance in SNNs, a pre-trained ANN acts as the teacher, providing robust feature extraction capabilities, while the SNN serves as the student, retaining its inherent advantages such as event-driven computation and low energy consumption\cite{yu_temporal_2025}. 
Distilling Spikes\cite{kushawaha_distilling_2021} proposes a multi-stage distillation framework employing intermediate networks to improve student SNN performance. 
Spike-Thrift\cite{kundu_spike-thrift_2021} reduces spike activity via attention-guided ANN compression and sparse training. 
LaSNN\cite{hong_lasnn_2023} introduces a layer-wise distillation framework incorporating attention mechanisms to bridge the representation and transmission gap between ANNs and SNNs.
KDSNN\cite{xu_constructing_2023} presents a unified distillation strategy that combines both logits-based and feature-based knowledge transfer. 
SpikeBERT\cite{lv_spikebert_2023} adopts a two-phase distillation approach to transfer knowledge from BERT into a spiking language model. 
Joint A-SNN\cite{guo_joint_2023} integrates self-distillation and weight factorization to further enhance model performance. 
\cite{xu_reversing_2024} proposed an evolutionary method that combines knowledge distillation and pruning to optimize SNN structures.
The SAKD\cite{qiu_self-architectural_2024} applies a dual-phase training process, weight transfer and behavioral imitation. 
BKDSNN\cite{xu_bkdsnn_2025} framework enhances feature transferability through blurred knowledge distillation, yielding improved performance on complex datasets.
Despite these advancements, fully leveraging the spatio-temporal dynamics of spikes remains a significant challenge. Thus, the design of distillation strategies explicitly tailored for the unique characteristics of SNNs remains a key direction for future research, with the potential to achieve further improvements in both accuracy and energy efficiency.

\section{Methodology}

\begin{figure*}
    \centering
    \includegraphics[width=0.9\linewidth]{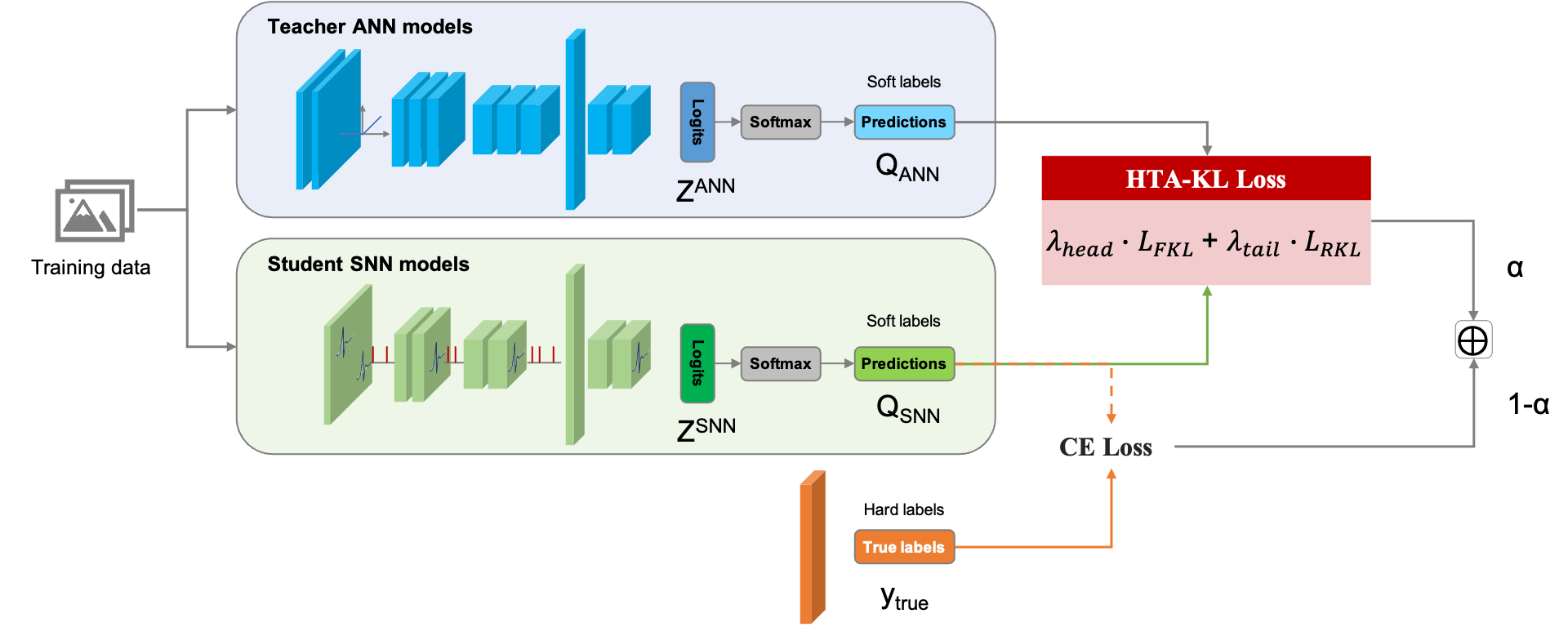}
    \caption{The framework of the proposed HTA-KL for SNN training.}
    \label{fig:archi}
\end{figure*}

In this section, we introduce HTA-KL, a novel knowledge distillation method specifically designed for SNNs. We first outline the preliminaries of SNNs and knowledge distillation, and then describe the HTA-KL divergence loss, a hybrid loss function we propose for more effective SNN distillation.

\subsection{Preliminaries}

\paragraph{Leaky Integrate-and-Fire Model}\label{sec:LIF}
The Leaky Integrate-and-Fire (LIF) neuron is a widely used model in SNNs due to its biological plausibility and computational efficiency. 
It can be mathematically represented as:

\begin{subequations}
    \begin{align}
    V^{t,n} &= f(H^{t-1,n}, X^{t,n}), \label{equ:charging} \\
    S^{t,n} &= \Theta(V^{t,n} - v_{\text{th}}), \label{equ:firing}\\
    H^{t,n} &= V_{\text{reset}} \cdot S^{t,n} + V^{t,n} \odot (1 - S^{t,n}). \label{equ:resetting}
\end{align}
\end{subequations}
where \(H^{t-1,n}\) represents the membrane potential after a spike trigger at the previous timestep, and \(X^{t,n}\) is the input feature of the \(n\)-th layer at timestep \(t\). The integrated membrane potential is denoted as \(V^{t,n}\), and the firing threshold \(v_{\text{th}}\) determines whether a spike \(S^{t,n}\) is triggered.
The Heaviside step function \(\Theta(x)\) is defined as \(\Theta(x) = 0\) if \(x < 0\), and \(\Theta(x) = 1\) if \(x \geq 0\).

%

\paragraph{Knowledge Distillation for SNNs}

Given a \(C\)-class classification task, let \(Z \in \mathbb{R}^{C}\) be the output logits of the network at each timestep \(t\), and \(Q\) represents the class probabilities. The class probability at timestep \(t\) is:

\begin{equation}
    Q_i = \frac{\exp(Z_i / \tau)}{\sum_{j=1}^{C}{\exp(Z_j / \tau)}}.
\end{equation}
Where \(Q_i\) represents the predicted probability for the \(i\)-th class, $Z_i$ represents the logits before softmax,
and \(\tau\) is the temperature scaling factor.
To minimize the difference between the teacher and student, the Kullback-Leibler (KL) divergence-based KD loss is computed as:

\begin{equation}
    \mathcal{L}_\text{KL}(Q^\text{ANN} || Q^\text{SNN}_{\text{avg}}) = \sum_{i=1}^{C} Q_{i}^\text{ANN} \log \left( \frac{Q_{i}^\text{ANN}}{Q_{i, \text{avg}}^\text{SNN}} \right).
\label{eq:kl}
\end{equation}
Where \( Q^\text{ANN} \) and \( Q^\text{SNN} \) are the post-softmax outputs of the teacher ANN and student SNN models, respectively.
And \( Q^\text{SNN}_{\text{avg}} = \frac{1}{T} \sum_{t=1}^{T} Q^\text{SNN}(t) \), is a temporal average output across timesteps to stabilize the teacher-student mapping across timesteps.
Cross-entropy loss $\mathcal{L}_\text{CE}$ is used to ensure the student model learns the true distribution of the labels:

\begin{equation}
    \mathcal{L}_\text{CE} = \text{CrossEntropy}(Q^\text{SNN}, y_{true}) = -\sum_{i=1}^{C} y_{\text{true},i} \log Q_{i}^\text{SNN},
\label{eq:ce}
\end{equation}
where \(y_{true}\) is the ground truth labels.

The SNN KD loss can be formulated as:

\begin{equation}
    \mathcal{L}_\text{SKD} = (1-\alpha) \cdot \mathcal{L}_\text{CE} + \alpha \cdot \mathcal{L}_\text{KL},
\label{eq:kd}
\end{equation}
\(\alpha\) is the loss weight.

\paragraph{FKL and RKL}

Forward Kullback-Leibler (FKL) and Reverse Kullback-Leibler (RKL) are two primary forms of Kullback-Leibler divergence used in KD. 
Both measure the divergence between the teacher's and student's distributions but focus on different aspects of the distribution.
Where FKL minimizes the divergence from the teacher’s distribution \( Q^\text{ANN} \) to the student’s distribution \( Q^\text{SNN} \):
\begin{equation}
    \mathcal{L}_\text{FKL} = \sum_{i=1}^{C} Q^\text{ANN}_i \log \left( \frac{Q^\text{ANN}_i}{Q^\text{SNN}_i} \right),
\end{equation}
encouraging the student to match the high-probability (head) regions of the teacher’s distribution. 
While RKL measures the divergence from the student’s distribution \( Q^\text{SNN} \) to the teacher’s \( Q^\text{ANN} \), focusing on the low-probability (tail) regions of the teacher's output:
\begin{equation}
    \mathcal{L}_\text{RKL} = \sum_{i=1}^{C} Q^\text{SNN}_i \log \left( \frac{Q^\text{SNN}_i}{Q^\text{ANN}_i} \right).
\end{equation}

\subsection{HTA-KL Divergence for SNNs}

Given the probability distributions \(Q^\text{ANN}\) and \(Q^\text{SNN}\) by applying the softmax to the logits as:

\begin{equation}
Q^\text{ANN}_i = \frac{\exp(Z^\text{ANN}i / \tau)}{\sum_{j=1}^{C} \exp(Z^\text{ANN}_j / \tau)},\\
Q^\text{SNN}_i = \frac{\exp(Z^\text{SNN}i / \tau)}{\sum_{j=1}^{C} \exp(Z^\text{SNN}_j / \tau)}.
\end{equation}
We sort the teacher’s probabilities \(Q^\text{ANN}\) in descending order:

\begin{equation}
\tilde{Q}^\text{ANN} = \text{sort}(Q^\text{ANN}),
\end{equation}
where \(\tilde{Q}^\text{ANN}\) is the sorted probabilities and \(\mathcal{I}\) denotes the corresponding sorting indices, which reorder \(Q^\text{SNN}\) accordingly:

\begin{equation}
\tilde{Q}^\text{SNN} = \text{gather}(Q^\text{SNN}, \mathcal{I}).
\end{equation}
This ensures that student sorted predictions are aligned with the teacher’s.
The absolute distance between the aligned distributions is computed as:

\begin{equation}
D_i = \left| \tilde{Q}^\text{ANN}_i - \tilde{Q}^\text{SNN}_i \right|,
\end{equation}
where \(D_i\) denotes the per-class divergence after alignment.
To differentiate between regions, we compute the cumulative sum over sorted teacher probabilities:

\begin{equation}
 S_i = \sum_{j=1}^{i} \tilde{Q}^\text{ANN}_j,
\end{equation}
where \(S\)  denotes the cumulative sum of sorted teacher probabilities.
Classes are divided into head and tail according to a threshold \(\delta\) , typically set to 0.5:

\begin{align}
\mathcal{M}_{\text{head}}(i) &=
\begin{cases}
1, & \text{if } S_i < \delta, \\
0, & \text{otherwise},
\end{cases}
\\
\mathcal{M}_{\text{tail}}(i) &= 1 - \mathcal{M}_{\text{head}}(i).
\end{align}
The head and tail distances are then defined as
\(d_{\text{head}} = \sum_{i} D_i \cdot \mathcal{M}_{\text{head}}(i)\) and \(d_{\text{tail}} = \sum_{i} D_i \cdot \mathcal{M}_{\text{tail}}(i)\).
Then the corresponding region-specific weights are:

\begin{equation}
    \lambda_{\text{head}} = \frac{d_{\text{head}}}{d_{\text{head}} + d_{\text{tail}}} \quad 
    \lambda_{\text{tail}} = \frac{d_{\text{tail}}}{d_{\text{head}} + d_{\text{tail}}}.
\end{equation}
The HTA-KL loss is computed as a weighted combination of FKL and RKL:

\begin{equation}
    \mathcal{L}_\text{HTA-KL} = \lambda_\text{head} \cdot \mathcal{L}_\text{FKL} + \lambda_\text{tail} \cdot \mathcal{L}_\text{RKL}.
\end{equation}

HTA-KL integrates FKL and RKL through a dynamic weighting scheme that adjusts their relative contributions based on the distributional characteristics of the teacher and the learning dynamics of the student. 
Specifically, FKL emphasizes alignment in the high-probability regions of the teacher’s output, while RKL focuses on the low-probability regions, encouraging the student to learn tail distributed patterns. 
The adaptive weighting assigns different importance to these regions based on the teacher's distribution and the student's performance.



\section{Experiments}
In this section, we present extensive experimental evaluations to demonstrate the effectiveness of HTA-KL compared to existing SNN distillation methods. 

\begin{table*}[h]
    \centering
    \begin{small}
    \caption{Comparison results with training-based SNN SOTA methods, including CNN-based and transformer-based approaches, on CIFAR-10 and CIFAR-100, with and without KD using ResNet19, ResNet20, and VGG16.}
    \begin{tabular}{lllllll}
        \toprule
        & \multirow{2}{*}{\textbf{Methods}} & \multirow{2}{*}{\textbf{Architecture}} & \multicolumn{4}{c}{\textbf{Accuracy(CIFAR-10 / CIFAR-100)}}  \\
        \cmidrule(lr){4-7}
       & &  & Timestep=1 & Timestep=2 & Timestep=4 & Timestep=6  \\
        \midrule
        \multirow{16}{*}{\rotatebox{90}{\textbf{without KD}}}
 & RecDis-SNN\cite{guo_recdis-snn_2022}    & ResNet-19 & - & - & 95.53\% / 74.10\% & - \\
\cmidrule{2-7}
& TET\cite{deng_temporal_2021} & ResNet-19 & - & 94.16\% / 72.87\% & 94.44\% / 74.47\% & 94.50\% / 74.72\% \\
\cmidrule{2-7}
&LSG\cite{lian_learnable_2023} & ResNet-19 & - & 94.41\% / 76.32\% & 95.17\% / 76.85\% & 95.52\% / 77.13\%\\
\cmidrule{2-7}
&PFA\cite{deng_tensor_2023}  & ResNet-19 & - & 95.60\% / 76.70\% & 95.71\% / 78.10\% & 95.70\% / 79.10\%  \\
\cmidrule{2-7}
&IM-loss\cite{guo_im-loss_2022}  & ResNet-19 & - & 93.85\% / - & 95.40\% / -& 95.49\% / -\\
\cmidrule{2-7}
&\multirow{2}{*}{MPBN\cite{guo_membrane_2023}} & ResNet-19  & 96.06\% / 78.71\% & 96.47\% / 79.51\% & 96.52\% / 80.10\% & - \\
                        &  & ResNet-20 & 92.22\%  /  - & 93.54\% / 70.79\% & 94.28\% / 72.30\% & - \\
\cmidrule{2-7}
&IM-LIF\cite{lian_im-lif_2024}  & ResNet-19 & - & 95.29\% / 77.21\% & - &  95.66\% / 77.42\%  \\      
\cmidrule{2-7}
&\multirow{2}{*}{Spikformer\cite{zhou_spikformer_2022}} & Spikformer-4-256 & - & - & 93.94\% / 75.96\% & - \\ & &  Spikformer-4-384 & - & - & 95.19\% / 77.86\% & - \\
\cmidrule{2-7}
& \multirow{2}{*}{Spikingformer\cite{zhou_spikingformer_2023}} & Spikingformer-4-256 & - & - & 94.77\% / 77.43\% & - \\ & &  Spikingformer-4-384 & - & - & 95.61\% / 79.36\% & - \\
\cmidrule{2-7}
& STMixer\cite{deng_spiking_2024} & STMixer-4-384-32 & 95.49\% / \textbf{80.00}\% & - & 96.01\% / \textbf{81.87}\% & - \\ 
\midrule
\multirow{16}{*}{\rotatebox{90}{\textbf{with KD}}} 
& \multirow{3}{*}{Teacher ANN} & ResNet19 &  97.51\% / 81.99\% & - & -& -\\
                &             & ResNet20  &  95.72\% / 75.37\% & - & -& -\\
                   &             & VGG16  &  96.22\% / 77.36\% & - & -& -\\
\cmidrule{2-7}
       & \multirow{3}{*}{KDSNN\cite{xu_constructing_2023}}
         & ResNet19 & 95.51\% / 78.25\% & 95.98\% / 79.45\% & 96.28\% / 80.06\% & -\\
       & & ResNet20 & - & 93.05\% / 70.85\% & 94.07\% / 72.41\% & 94.17\% / 73.29\%  \\
       & & VGG16 & - & 94.50\% / 73.48\% & 94.88\% / 74.75\% & 94.83\% / 75.17\%  \\
\cmidrule{2-7}
       & \multirow{3}{*}{LASNN\cite{hong_lasnn_2023}}
         & ResNet19 & 96.19\% / 78.69\% & 96.54\% / 79.52\% & 96.60\% / 80.22\% & - \\
       & & ResNet20 & - & 92.90\% / 68.63\% & 93.36\% / 69.96\% & 93.77\% / 70.77\%  \\
       & & VGG16    & - & 94.27\% / 73.15\% & 94.64\% / 73.86\% & 94.69\% / 74.29\% \\
\cmidrule{2-7}
        &\multirow{3}{*}{BKDSNN\cite{xu_bkdsnn_2025}} 
         & ResNet19 & 96.03\% / 78.77\% & 96.30\% / 79.98\% & 96.70\% / 80.64\% & - \\
        && ResNet20 & - & 93.09\% / 71.21\% & 94.02\% / 72.17\% & 94.17\% / 73.20\%  \\
        && VGG16    & - & 94.61\% / 73.30\% & 94.89\% / 74.92\% & 94.87\% / 74.64\%  \\
\cmidrule{2-7}
         & \multirow{3}{*}{\textbf{Ours HTA-KL}} 
  & ResNet19 & \textbf{96.11}\% / 78.75\% & \textbf{96.68}\% / \textbf{80.51}\% & \textbf{96.76}\% / 81.03\% & - \\ 
& & ResNet20 &  - & 93.38\% / 71.89\% & 94.06\% / 73.98\% & 94.31\% / 74.28\% \\
& & VGG16    & - & 94.44\% / 74.64\% & 94.93\% / 75.88\% &  \textbf{94.96}\% / \textbf{76.02}\%  \\
        \bottomrule
    \end{tabular}
    \end{small}
    \label{tab:results}
\end{table*}

\subsection{Experimental Settings}
\label{sec:setting}

\paragraph{Datasets.}

We evaluate HTA-KL on three widely used image classification datasets: CIFAR-10, CIFAR-100 \cite{krizhevsky_cifar-10_2010} and Tiny ImageNet\cite{le_tiny_2015}. 
CIFAR-10 contains 50,000 training and 10,000 testing images, each of size $32 \times 32$, across 10 classes, while CIFAR-100 contains 100 classes with the same image size and split. 
Additionally, we conduct experiments on Tiny ImageNet, which consists of 100 classes with 500 training images per class and 50 validation images per class, each resized to $64 \times 64$ pixels.
Standard data augmentation techniques, such as random cropping and horizontal flipping, are applied, with the datasets split in a 9:1 training/testing ratio.
Our results on Tiny ImageNet further demonstrate HTA-KL’s scalability and effectiveness on more complex datasets.

\paragraph{Implementation Details.}
All experiments were conducted on a high-performance server with 8 NVIDIA RTX 4090 GPUs and a 64-core AMD EPYC 7763 CPU, using PyTorch and SpikingJelly for model implementation. We use ResNet-19, ResNet-20, and VGG-16 as student models, with corresponding ANN counterparts as teacher models. The training schedule spanned 300 epochs, with an initial learning rate of 0.01, and optimization was performed using SGD with momentum (0.9) and a batch size of 128. We test timesteps of 1, 2, 4, and 6 to evaluate the trade-off between latency and accuracy. Standard data augmentation techniques (random cropping, horizontal flipping) were applied.

\subsection{Comparison with Knowledge distillation Methods}
\label{sec:sota}

\paragraph{CIFAR-10 and CIFAR-100} 
We compared HTA-KL against three existing SNN distillation methods: KDSNN, LaSNN, and BKDSNN, as shown in \cref{tab:results}. 
Results on CIFAR-10 and CIFAR-100 demonstrate that HTA-KL consistently outperforms these methods across all architectures and timesteps. 
For example, on CIFAR-100, HTA-KL improves ResNet-19 accuracy by 0.53\%, 0.39\%, and 0.85\% at timesteps 2, 4, and 6 with other KD methods, respectively. 
HTA-KL requires fewer inference timesteps compared to prior works, making it a more practical solution for energy-efficient SNNs.

\begin{figure}[t]
    \centering
    \includegraphics[width=0.9\linewidth]{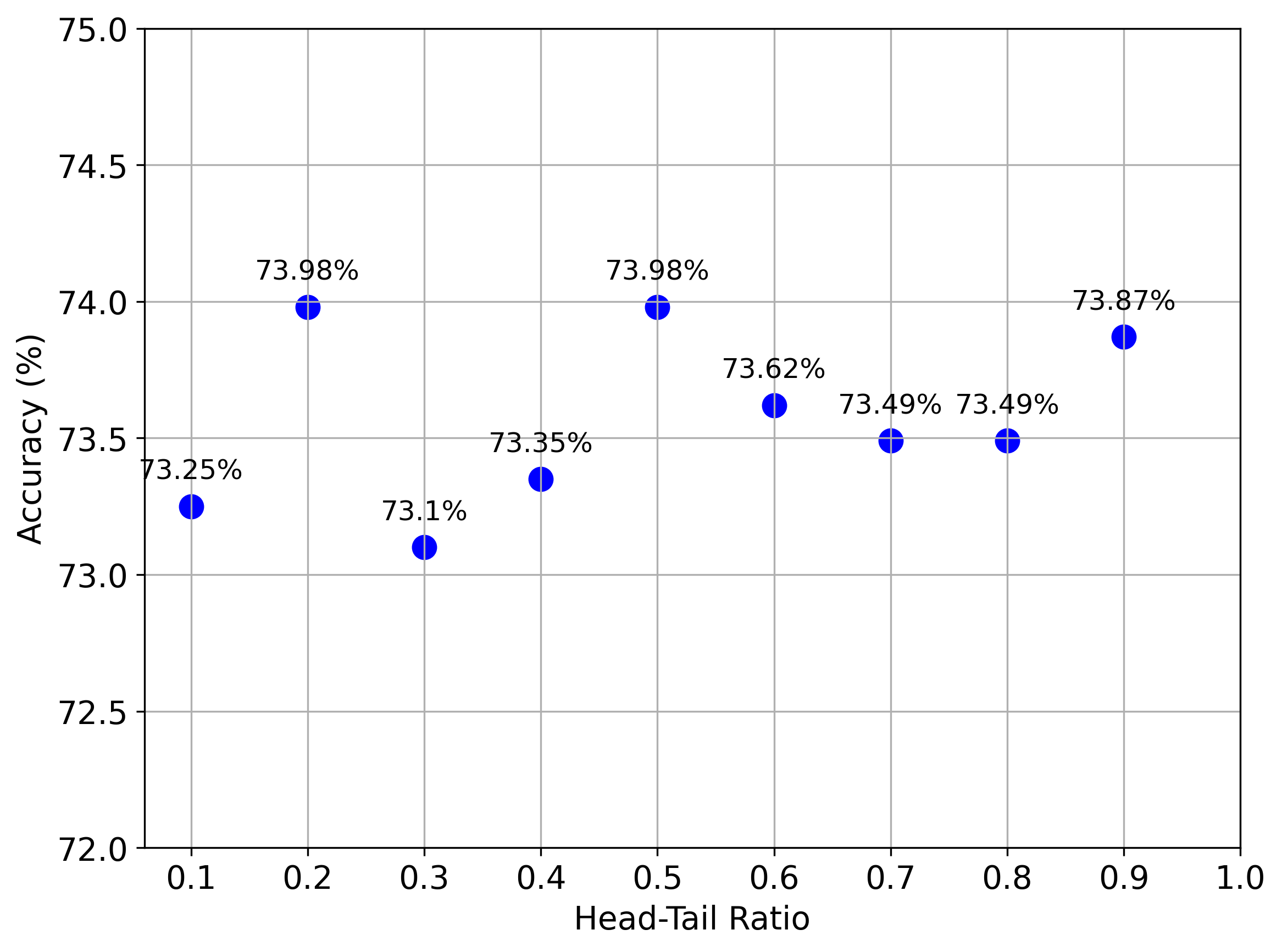}
        \caption{Impact of the Head-Tail Ratio on Accuracy. The graph shows the accuracy of ResNet-20 on CIFAR-100 at timestep 4, with different ratios between the head and tail losses. The accuracy improves as the ratio of the head loss increases, reaching optimal performance at a balanced head-tail ratio.}
    \label{fig:weight}
\end{figure}

\paragraph{Tiny-ImageNet}

\begin{table}[h]
\centering
\caption{Comparison results with training-based SNN SOTA methods, including CNN-based and transformer-based approaches, on Tiny-ImageNet, with and without KD. \textbf{T} denotes Timestep.}
\small
\begin{tabular}{lcccc}
\toprule 
& \textbf{Method}                         & \textbf{Architecture}       & \textbf{T}   & \textbf{Accuracy}         \\ 
\midrule
\multirow{10}{*}{\rotatebox{90}{\textbf{without KD}}} 
& TET~\cite{deng_temporal_2021}                           & R-34          & 6          & 64.79\%          \\
\cmidrule{2-5}
& GLIF~\cite{yao_glif_2022}                           & R-34          & 4          & 67.52\%          \\
\cmidrule{2-5}
& \multirow{2}{*}{MPBN~\cite{guo_membrane_2023}}          & R-18          & 4          & 63.14\%          \\
&                               & R-34          & 4          & 64.71\%          \\ 
\cmidrule{2-5}
& \multirow{2}{*}{SEW ResNet~\cite{fang_deep_2021}}    & R-18          & 4          & 63.18\%          \\
&                               & R-34          & 4          & 67.04\%          \\
\cmidrule{2-5}
& \multirow{2}{*}{Spikformer~\cite{zhou_spikformer_2022}} & Spikformer-8-384 & 4 & 70.24\% \\
&                               & Spkfmer-6-512          & 4          & 72.46\%          \\
\cmidrule{2-5}
& Spikingformer~\cite{zhou_spikingformer_2023} & Spikingformer-8-384 & 4 & 72.45\%  \\ 
\midrule
\multirow{14}{*}{\rotatebox{90}{\textbf{with KD}}} 
& \multirow{2}{*}{Teacher ANN} & ResNet-20 & 1 & 65.72\% \\
&             & VGG-16 & 1  & 65.94\% \\
\midrule
& \multirow{2}{*}{LaSNN~\cite{hong_lasnn_2023}}                 & SEW R-18    & 4    & 63.33\%    \\
&                      & SEW R-34    & 4   & 66.98\%   \\ 
\cmidrule{2-5}
& \multirow{2}{*}{KDSNN~\cite{xu_constructing_2023}}                 & SEW R-18    & 4    & 63.61\%   \\
&                      & SEW R-34    & 4    & 67.28\%  \\ 
\cmidrule{2-5}
& \multirow{2}{*}{BKDSNN~\cite{xu_bkdsnn_2025}}                & SEW R-18    & 4    & 63.43\%  \\
&                      & SEW R-34   & 4    & 67.21\%  \\ 
\cmidrule{2-5}
& \multirow{4}{*}{\textbf{Ours}} & \multirow{2}{*}{ResNet-20} & 1 & 61.28\%  \\
 & & & 2 & 64.32\% \\
 \cmidrule{3-5}
&                               & \multirow{2}{*}{VGG-16} & 1 & 62.60\% \\ 
 & & & 2 & 64.10\%  \\
\bottomrule
\end{tabular}
\label{tab:imagenet}
\end{table}

\begin{table*}[h]
    \centering
    \caption{Comparison of energy consumption (mJ), spike firing rates (\%), and computational operations (ACs and MACs) across different SNN distillation methods. 
    }
    \begin{tabular}{llccccc}
        \toprule
        \textbf{Methods} & \textbf{Architecture} & \textbf{Firing Rate} & \textbf{OPs} & \textbf{ACs} & \textbf{MACs} & \textbf{Energy} \\
        \midrule
        \multirow{4}{*}{ResNet19} & KDSNN & 29.96\% & 9.15G & 2.21G & 26.68M & 2.11173mJ \\
        & LASNN & 36.49\% & 9.15G & 2.78G & 26.68M & 2.62473mJ \\
        & BKDSNN & 21.55\% & 9.15G & 1.85G & 26.68M & 1.78773mJ \\
        & \textbf{Ours HTA-KL} & 27.44\% & 9.15G & 2.11G & 26.68M & 2.02173mJ \\
        \midrule
        \multirow{4}{*}{ResNet20} & KDSNN & 29.35\% & 865.98M & 213.25M & 53.99M & 0.440279mJ \\
        & LASNN & 34.51\% & 865.98M & 238.88M & 53.99M & 0.463346mJ \\
        & BKDSNN & 29.42\% & 865.98M & 214.3M & 53.99M & 0.441224mJ \\
        & \textbf{Ours HTA-KL} & 27.91\% & 865.98M & 198.2M & 53.99M & 0.426734mJ \\
        \midrule
        \multirow{4}{*}{VGG16} & KDSNN & 22.55\% & 1.26G & 196.63M & 274.24M & 1.438471mJ \\
        & LASNN & 26.13\% & 1.26G & 217.22M & 274.24M & 1.457002mJ \\
        & BKDSNN & 23.69\% & 1.26G & 202.53M & 274.24M & 1.443781mJ \\
        & \textbf{Ours HTA-KL} & 23.42\% & 1.26G & 199.39M & 274.24M & 1.440955mJ \\
        \bottomrule
    \end{tabular}
    \label{tab:energy}
\end{table*}

For Tiny ImageNet, we integrated HTA-KL into ResNet-20 and VGG-16. As shown in \cref{tab:imagenet}, ResNet-20 achieves 64.32\% accuracy at timestep 2, comparable to methods that use timestep 4. This demonstrates that HTA-KL achieves performance with fewer timesteps, reducing computational cost and energy consumption. 

\begin{figure}
    \centering
    \includegraphics[width=0.9\linewidth]{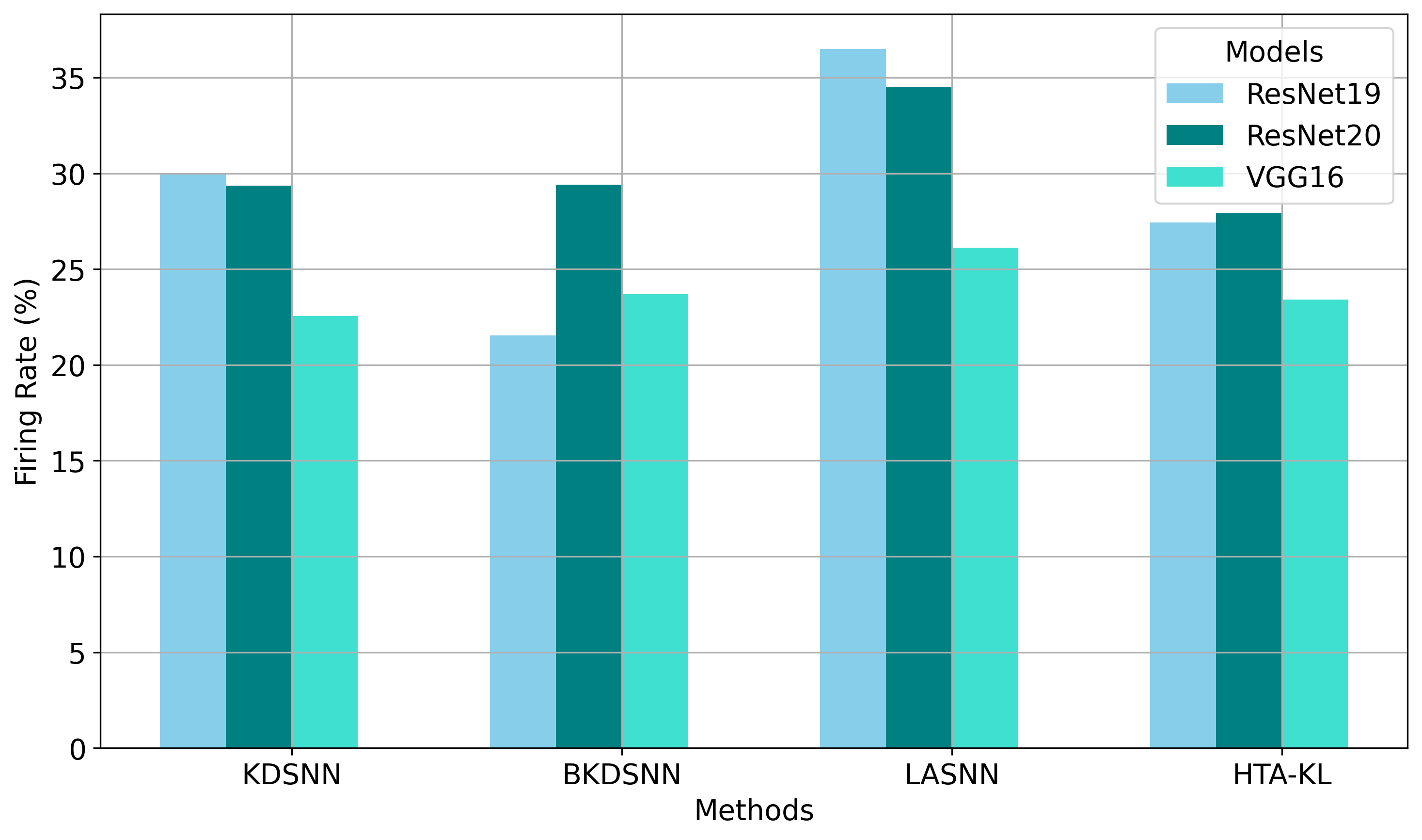}
    \caption{Spike firing rate comparison for different distillation methods on CIFAR-100. The figure shows the firing rate across different models (ResNet19, ResNet20, and VGG16) and methods (KDSNN, BKDSNN, LASNN, and HTA-KL).}
    \label{fig:firing}
\end{figure}

\begin{figure*}[!h]
    \centering
    \includegraphics[width=0.9\linewidth]{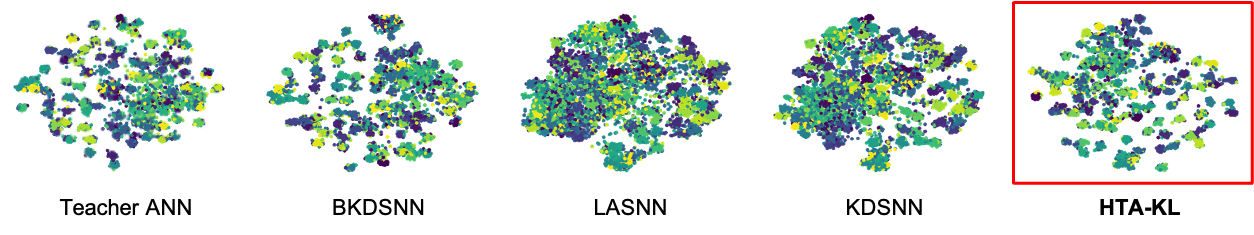}
    \caption{t-SNE Visualization of features learned by teacher ANN and different distillation methods.}
    \label{fig:visual}
\end{figure*}

\subsection{Impact of FKL and RKL Ratio on Model Accuracy}
\label{sec:exp_ablation}

To optimize HTA-KL method and better understand its effectiveness, we conducted ablation studies to analyze the impact of the ratio between the head and tail losses on model accuracy. 
As shown in \cref{fig:weight}, varying the head-tail ratio significantly affects the accuracy of the model. Our results demonstrate that the accuracy improves as the weight given to the head loss increases, with the best performance observed at a balanced head-tail ratio. This suggests that an optimal balance between the head and tail losses is crucial for improving training stability and enhancing knowledge transfer from the teacher model to the student. 
The stability of the model further emphasizes the importance of adaptive weighting in HTA-KL to achieve high accuracy and efficient knowledge distillation.

\subsection{Spike Firing Rate Analysis.}
\label{sec:perf}

We evaluate the spike firing rate of intermediate feature maps before the last layer in ResNet19, ResNet20, and VGG16 across different distillation methods on the CIFAR-100 dataset. 
The spike firing rate reflects the proportion of active spikes in the network, which is a crucial factor in understanding computational efficiency.

As shown in \cref{fig:firing}, our method, HTA-KL, achieves a competitive balance between firing rate and efficiency. Specifically, HTA-KL maintains a moderate firing rate across all models, outperforming KDSNN, BKDSNN, and LASNN in terms of efficiency while achieving higher accuracy. For example, ResNet19 with HTA-KL achieves a firing rate of 27.44\%, which is lower than that of LASNN (36.49\%) but still ensures effective model performance. Similar trends are observed in ResNet20 and VGG16, where HTA-KL consistently provides a more balanced firing rate compared to other methods. 
These results highlight that controlling the spike firing rate is crucial for optimizing computational efficiency without compromising performance. By reducing unnecessary spikes, HTA-KL enables more efficient processing, making it well-suited for energy-constrained environments while maintaining strong performance across different architectures.

\subsection{Energy Consumption Analysis.}\label{sec:energy}
We evaluate the energy efficiency of HTA-KL by estimating energy consumption based on the number of operations (OPs) in 45nm technology during single-image inference. Unlike ANNs, where MACs remain constant, SNN computations depend on ACs, which are dynamically triggered by spikes. Using the model from \cite{horowitz_11_2014, qiao_reconfigurable_2015}, MACs consume 4.6 pJ per operation, while ACs require only 0.9 pJ, highlighting the energy efficiency of SNNs.

As shown in \cref{tab:energy}, HTA-KL achieves a favorable trade-off between accuracy and computational cost. Specifically, ResNet-19 with HTA-KL consumes 2.02173mJ, outperforming KDSNN and BKDSNN in energy efficiency while maintaining a competitive firing rate. Similar trends are observed for ResNet-20 and VGG-16, confirming HTA-KL’s ability to balance performance and energy efficiency, making it suitable for real-world applications where both accuracy and energy conservation are critical.

\subsection{Visualization.}\label{sec:visual}

We use t-SNE to visualize the features learned by different distillation methods on the CIFAR-100 dataset, with ResNet-20 as the teacher model and its corresponding SNN as the student. As shown in \cref{fig:visual}, HTA-KL achieves better feature distinguishability compared to previous SNN distillation techniques. The t-SNE plots demonstrate that HTA-KL leads to more separable and well-structured features, especially in deeper layers, highlighting its effectiveness in transferring knowledge from the teacher to the student model. Other methods show more overlapping features, indicating less effective feature learning.

\section{Conclusion}
In this paper, we introduced Head-Tail Aware HTA-KL divergence, a novel knowledge distillation method designed specifically for SNNs. 
HTA-KL dynamically distinguishes between high- and low-probability regions, using adaptive weights to balance the importance of both head and tail distribution. Our experimental results demonstrate that HTA-KL significantly improves SNN performance, bridging the gap between SNNs and traditional ANNs while maintaining energy efficiency and biological plausibility. This approach offers a promising solution for advancing SNNs in various applications, and future work can explore its optimization, adaptation to different network architectures, and performance on more complex datasets.

\section*{Acknowledgment}

This work is supported by Zhejiang Provincial Natural Science Foundation of China (LDT23F02023F02) and Research Fund for International Scientists of National Natural Science Foundation of China (72350710798).

\bibliographystyle{IEEEtran}
\bibliography{references}

\end{document}